\pdfoutput=1

\documentclass[11pt,a4paper]{article}
\usepackage[final]{naacl2021}
\usepackage{times}
\usepackage{latexsym}
\usepackage{graphics}
\usepackage{graphicx}
\usepackage{enumitem}
\usepackage{amsmath}
\usepackage{amssymb}
\usepackage{subcaption}
\usepackage[labelfont=bf]{caption}

\usepackage{booktabs}
\usepackage[utf8]{inputenc}
\usepackage{url}

\usepackage{microtype}

\title{On the State of Social Media Data for Mental Health Research}

\author{Keith Harrigian, Carlos Aguirre, Mark Dredze \\
  Johns Hopkins University \\
  \texttt{kharrigian@jhu.edu, caguirr4@jhu.edu, mdredze@cs.jhu.edu}}

\date{}

\begin{document}

\maketitle


\begin{abstract}
Data-driven methods for mental health treatment and surveillance have become a major focus in computational science research in the last decade. However, progress in the domain remains bounded by the availability of adequate data. Prior systematic reviews have not necessarily made it possible to measure the degree to which data-related challenges have affected research progress. In this paper, we offer an analysis specifically on the state of social media data that exists for conducting mental health research. We do so by introducing an open-source directory of mental health datasets, annotated using a standardized schema to facilitate meta-analysis.\footnote{\url{https://github.com/kharrigian/mental-health-datasets}} 

\end{abstract}


\section{Introduction}

The last decade has seen exponential growth in computational research devoted to modeling mental health phenomena using non-clinical data \cite{bucci2019digital}. Studies analyzing data from the web, such as social media platforms and peer-to-peer messaging services, have been particularly appealing to the research community due to their scale and deep entrenchment within contemporary culture \cite{perrin2015social,fuchs2015culture,graham2015role}. Such studies have yielded novel insights into population-level mental health \cite{de2013predicting,amir-etal-2019-mental} and shown promising avenues for the incorporation of data-driven analyses in the treatment of psychiatric disorders \cite{eichstaedt2018facebook}.

These research achievements have come despite complexities specific to the mental health space often making it difficult to obtain a sufficient sample size of high-quality data. For instance, behavioral disorders are known to display variable clinical presentations amongst different populations, rendering annotations of ground truth inherently noisy \cite{Choudhury2017GenderAC,arseniev2018type}. Scalable methods for capturing an individual's mental health status, such as using regular expressions to identify self-reported diagnoses or grouping individuals based on activity patterns, have provided opportunities to construct datasets aware of this heterogeneity \cite{coppersmith-etal-2015-clpsych,Kumar2015DetectingCI}. However, they typically rely on oversimplifications that lack the same clinical validation and robustness as something like a mental health battery \cite{Zhang2014UsingLF,ernala2019methodological}.

Ethical considerations further complicate data acquisition, with the sensitive nature of mental health data requiring tremendous care when constructing, analyzing, and sharing datasets \cite{benton2017ethical}. Privacy-preserving measures, such as de-identifying individuals and requiring IRB approval to access data, have made it possible to share some data across research groups. However, these mechanisms can be technically cumbersome to implement and are subject to strict governance policies when clinical information is involved due to HIPAA \cite{price2019privacy}. Moreover, many privacy-preserving practices require that signal relevant to modeling mental health, such as an individual's demographics or their social network, are discarded \cite{bakken2004data}. This missingness has the potential to limit algorithmic fairness, statistical generalizability, and experimental reproducibility \cite{gorelick2006bias}. Although mental health researchers may anecdotally recall difficulties acquiring quality data or reproducing prior art due to data sharing constraints, no study to our knowledge has explicitly quantified this challenge.


Indeed, prior reviews of computational research for mental health have noted several of the aforementioned challenges, but have predominantly discussed technical methods (e.g. model architectures, feature engineering) developed to surmount existing constraints \cite{guntuku2017detecting,wongkoblap2017researching}. Recent work from \citet{chancellor2020methods}, completed concurrently with our own, was the first review to focus specifically on the shortcomings of \emph{data} for mental health research. Our study affirms the findings of \citet{chancellor2020methods}, using an expanded pool of literature that more acutely focuses on \emph{language} found in social media data. To this end, we construct a new open-source directory of mental health datasets, annotated using a standardized schema that not only enables researchers to identify \emph{relevant} datasets, but also to identify \emph{accessible} datasets. We draw upon this resource to offer nuanced recommendations regarding future dataset curation efforts.



\section{Data}

To generate evidence-based recommendations regarding mental health dataset curation, we require knowledge of the extant data landscape. Unlike some computational fields which have a surplus of well-defined and uniformly-adopted benchmark datasets, mental health researchers have thus far relied on a decentralized medley of resources. This fact, spurred in part by the variable presentations of psychiatric conditions and in part by the sensitive nature of mental health data, thus requires us to compile a new database of literature. In this section, we detail our literature search, establish inclusion/exclusion criteria, and define a list of dataset attributes to analyze.

\subsection{Dataset Identification}

Datasets were sourced using a breadth-focused literature search. After including data sources from the three aforementioned systematic reviews \cite{guntuku2017detecting,wongkoblap2017researching,chancellor2020methods}, we searched for literature that lie primarily at the intersection of natural language processing (NLP) and mental health communities. We sought peer-reviewed studies published between January 2012 and December 2019 in relevant conferences (e.g. NAACL, EMNLP, ACL, COLING), workshops (e.g. CLPsych, LOUHI), and health-focused journals (e.g. JMIR, PNAS, BMJ).

We searched Google Scholar, ArXiv, and PubMed to identify additional candidate articles. We used two search term structures --- 1) (mental health $\vert$ \texttt{DISORDER}) + (social $\vert$ electronic) + media, and 2) (machine learning $\vert$ prediction $\vert$ inference $\vert$ detection) + (mental health $\vert$ \texttt{DISORDER}). `$\vert$' indicates a logical or, and \texttt{DISORDER} was replaced by one of 13 mental health keywords.\footnote{Depression, Suicide, Anxiety, Mood, PTSD, Bipolar, Borderline Personality, ADHD, OCD, Panic, Addiction, Eating, Schizophrenia} Additional literature was identified using snowball sampling from the citations of these papers. To moderately restrict the scope of this work, computational research regarding neurodegenerative disorders (e.g. Dementia, Parkinson's Disease) was ignored.

\subsection{Selection Criteria}

To enhance parity amongst datasets considered in our meta-analysis, we require datasets found within the literature search to meet three additional criteria. While excluded from subsequent analysis, datasets that do not meet this criteria are maintained with complete annotations in the aforementioned digital directory. In future work, we will expand our scope of analysis to reflect the multi-faceted computational approaches used by the research community to understand mental health.


\begin{enumerate}
    \itemsep0em
    \item Datasets must contain non-clinical electronic media (e.g. social media, SMS, online forums, search query text).
    \item Datasets must contain written language (i.e. text) within each unit of data .
    \item Datasets must contain a dependent variable that captures or proxies a psychiatric condition listed in the DSM-5 \cite{american2013diagnostic}.
\end{enumerate}

Our first criteria excludes research that examines electronic health records or digitally-transcribed interviews \cite{Gratch2014TheDA,holderness-etal-2019-distinguishing}. Our second criteria excludes research that, for example, primarily analyzes search query volume or mobile activity traces \cite{ayers2013seasonality,renn2018smartphone}. It also excludes research based on speech data \cite{iter2018automatic}. Our third criteria excludes research in which annotations are only loosely associated with their stated mental health condition. For instance, we filter out research that seeks to identify diagnosis dates in self-disclosure statements \cite{MacAvaney2018RSDDTimeTA}, in addition to research that proposes using sentiment as a proxy for mental illness \cite{davcheva-adam-benlian-2019-user}. This last criteria also inherently excludes datasets that lack annotation of mental health status altogether (e.g. data dumps of online mental health support platforms and text-message counseling services) \cite{Loveys2018CrossculturalDI,demasi-etal-2019-towards}.

\subsection{Annotation Schema}

We develop a high-level schema to code properties of each dataset. In addition to standard reference information (i.e. Title, Year Published, Authors), we note the following characteristics: 


\begin{itemize}
    \itemsep0em 
    \item \textbf{Platforms}: Electronic media source (e.g. Twitter, SMS)
    \item \textbf{Tasks}: The mental health disorders included as dependent variables (e.g. depression, suicidal ideation, PTSD)
    \item \textbf{Annotation Method}: Method for defining and annotating mental health variables (e.g. regular expressions, community participation/affiliation, clinical diagnosis)
    \item \textbf{Annotation Level}: Resolution at which ground-truth annotations are made (e.g. individual, document, conversation)
    \item \textbf{Size}: Number of data points at each annotation resolution for each task class
    \item \textbf{Language}: The primary language of text in the dataset
    \item \textbf{Data Availability}: Whether the dataset can be shared and, if so, the mechanism by which it may be accessed (e.g. data usage agreement, reproducible via API, distribution prohibited by collection agreement)
\end{itemize}

If a characteristic is not clear from a dataset's associated literature, we leave the characteristic blank; missing data points are denoted where applicable. While we simplify these annotations for a standardized analysis --- e.g. different psychiatric batteries used to annotate depression in individuals (e.g. PHQ-9, CES-D) are simplified as ``Survey (Clinical)'' --- we maintain specifics in the digital directory.


\section{Analysis}

\begin{figure}[t!]
    \centering
    \includegraphics[width=\columnwidth]{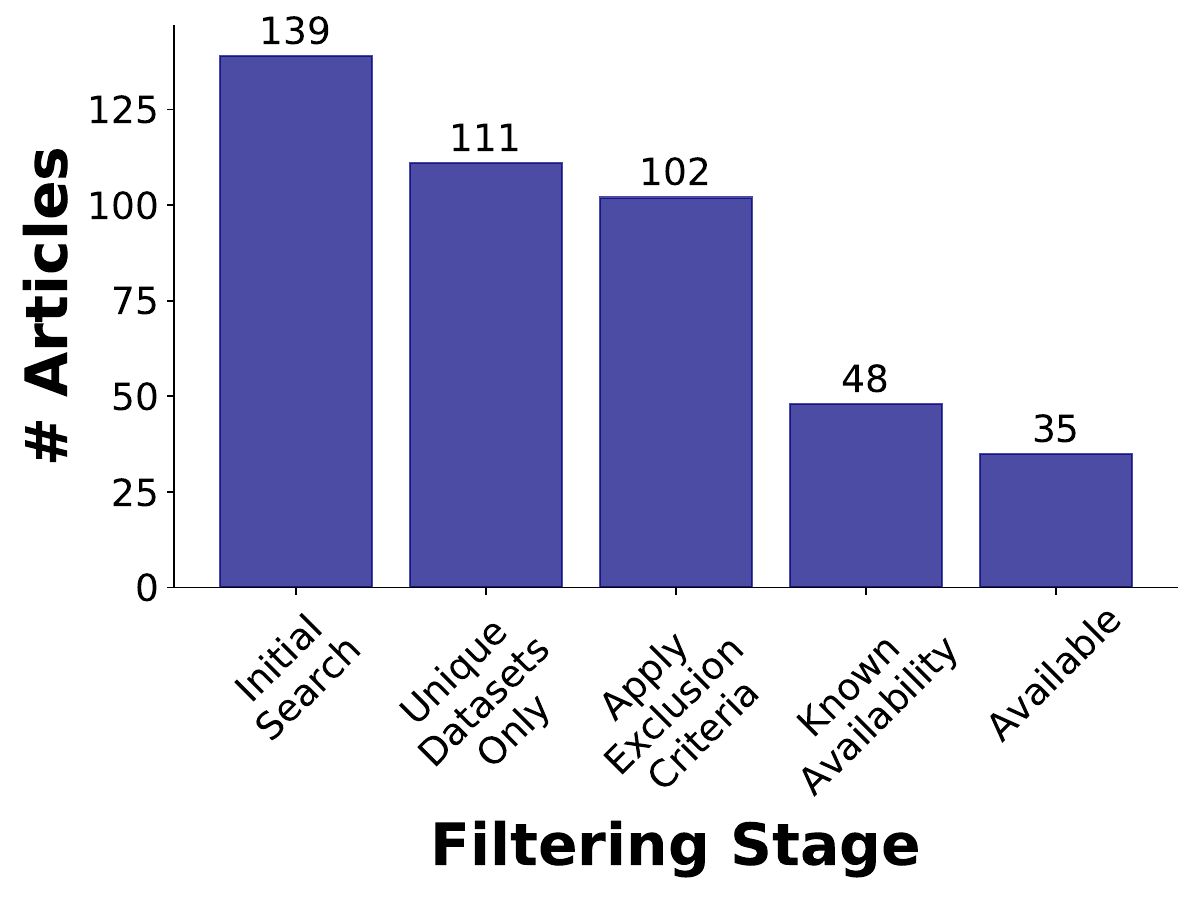}
    \caption{Number of articles (e.g. datasets) remaining after each stage of filtering. We were unable to readily discern the external availability of datasets for over half of the studies.}
    \label{fig:Search}
\end{figure}

Our literature search yielded 139 articles referencing 111 nominally-unique datasets. Application of exclusion criteria left us with 102 datasets. A majority of the datasets were released after 2012, with an average of 12.75 per year, a minimum of 1 (2012), and a maximum of 23 (2017). The 2015 CLPsych Shared Task \cite{coppersmith-etal-2015-clpsych}, Reddit Self-reported Depression Diagnosis \cite{Yates2017DepressionAS}, and ``Language of Mental Health'' \cite{gkotsis-etal-2016-language} datasets were the most reused resources, serving as the basis of 7, 3, and 3 additional publications respectively. All datasets known to be available for distribution are available with annotations in the appendix, while remaining datasets are found our digital directory.

\textbf{Platforms}. We identified 20 unique electronic media platforms across the 102 datasets. Twitter (47 datasets) and Reddit (22 datasets) were the most widely studied platforms. YouTube, Facebook, and Instagram were relatively underutilized for mental health research --- each found less than ten times in our analysis --- despite being the three most-widely adopted social media platforms globally \cite{perrin2019share}. We expect our focus on NLP to moderate the presence of YouTube and Instagram based datasets, though not entirely given both platforms offer expansive text fields (i.e. comments, tags) in addition to their primary content of video and images \cite{Chancellor2016QuantifyingAP,choi2016social}. It is more likely that use of these platforms (and Facebook) for research is hindered by increasingly stringent privacy policies and ethical concerns \cite{panger2016reassessing,benton2017ethical}.

\textbf{Tasks}. We identified 36 unique mental health related modeling tasks across the 102 datasets. While the majority of tasks were examined less than twice, a few tasks were considered quite frequently. Depression (42 datasets), suicidal ideation (26 datasets), and eating disorders (11 datasets) were the most common psychiatric conditions examined. Anxiety, PTSD, self-harm, bipolar disorder, and schizophrenia were also prominently featured conditions, each found within at least four unique datasets. A handful of studies sought to characterize finer-grained attributes associated with higher-level psychiatric conditions (e.g. symptoms of depression, stress events and stressor subjects) \cite{mowery-etal-2015-towards,lin2016does}. The dearth of anxiety-specific datasets was somewhat surprising given the condition's prevalence and the abundance of pyschometric batteries for assessing anxiety \cite{cougle2009anxiety,antony2020handbook}. That said, generalized anxiety disorder (GAD) only accounts for a small proportion of the overall prevalence of anxiety disorders \cite{bandelow2015epidemiology} and many other types of anxiety disorders (e.g. social anxiety, PTSD, OCD, etc.) were typically treated as independent conditions \cite{coppersmith-etal-2015-adhd,Choudhury2016DiscoveringST}.

\textbf{Annotation}. We identified 24 unique annotation mechanisms. It was common for several annotation mechanisms to be used jointly to increase precision of the defined task classes and/or evaluate the reliability of distantly supervised labeling processes. For example, some form of regular expression matching was used to construct 43 of datasets, with 23 of these including manual annotations as well. Community participation/affiliation (24 datasets), clinical surveys (22 datasets), and platform activity (3 datasets) were also common annotation mechanisms. The majority of datasets contained annotations made on the individual level (63 datasets), with the rest containing annotations made on the document level (40 datasets).\footnote{One dataset was annotated at both a document and individual level}

\textbf{Size}. Of the 63 datasets with individual-level annotations, 23 associated articles described the amount of documents and 62 noted the amount of individuals available. Of the 40 datasets with document-level annotations, 37 associated articles noted the amount of documents and 12 noted the number of unique individuals. The distribution of dataset sizes was primarily right-skewed.


One concerning trend that emerged across the datasets was the presence of a relatively low number of unique individuals. Indeed, these small sample sizes may further inhibit model generalization from platforms that are already demographically-skewed \cite{smith2018social}. The largest datasets, which present the strongest opportunity to mitigate the issues presented by poorly representative online populations, tend to leverage the noisiest annotation mechanisms. For example, datasets that define a mainstream online community as a control group may expect to find approximately 1 in 20 of the labeled individuals are actually living with mental health conditions such as depression \cite{wolohan-etal-2018-detecting}, while regular expressions may fail to distinguish between true and non-genuine disclosures of a mental health disorder up to 10\% of the time \cite{Cohan2018SMHDAL}.

\textbf{Primary Language}. Six primary languages were found amongst the 102 datasets --- English (85 datasets), Chinese (10 datasets), Japanese (4 datasets), Korean (2 datasets), Spanish (1 dataset), and Portuguese (1 dataset). This is not to say that some of the datasets do not include other languages, but rather that the predominant language found in the datasets occurs with this distribution. While an overwhelming focus on English data is a theme throughout the NLP community, it is a specific concern in this domain where culture often influences the presentation of mental health disorders \cite{Choudhury2017GenderAC,Loveys2018CrossculturalDI}.

\textbf{Availability}. We were able to identify the availability of only 48 of the 102 unique datasets in our literature search. Of these 48 datasets, 13 were known not to be available for distribution, generally due to limitations defined in the original collection agreement or removal from the public record \cite{Park2012DepressiveMO,schwartz-etal-2014-towards}. The remaining 35 datasets were available via the following distribution mechanisms: 18 may be reproduced using an API and instructions provided within the associated article, 12 require a signed data usage agreement and/or IRB approval, 3 are available without restriction, and 2 may be retrieved directly from the author(s) with permission. Of the 22 datasets that used clinically-derived annotations (e.g. mental health battery, medical history), 7 were unavailable for distribution due to terms of the original data collection process and 1 was removed from the public record. The remaining 14 had unknown availability.


\section{Discussion}

In this study, we introduced and analyzed a standardized directory of social media datasets used by computational scientists to model mental health phenomena. In doing so, we have provided a valuable resource poised to help researchers quickly identify new datasets that support novel research. Moreover, we have provided evidence that affirms conclusions from \citet{chancellor2020methods} and may further encourage researchers to rectify existing gaps in the data landscape. Based on this evidence, we will now discuss potential areas of improvement within the field.

\textbf{Unifying Task Definitions}. In just 102 datasets, we identified 24 unique annotation mechanisms used to label over 35 types of mental health phenomena. This total represents a conservative estimate given that nominally equivalent annotation procedures often varied non-trivially between datasets (e.g. PHQ-9 vs. CES-D assessments, affiliations based on Twitter followers vs. engagement with a subreddit) \cite{faravelli1986assessment,pirina2018identifying}. Minor discrepancies in task definition reflect the heterogeneity of how several mental health conditions manifest, but also introduce difficulty contextualizing results between different studies. Moreover, many of these definitions may still fall short of capturing the nuances of mental health disorders \cite{arseniev2018type}. As researchers look to transition computational models into the clinical setting, it is imperative they have access to standardized benchmarks that inform interpretation of predictive results in a consistent manner \cite{norgeot2020minimum}.

\textbf{Sharing Sensitive Data}. Most existing mental health datasets rely on some form of self-reporting or distinctive behavior to assign individuals into task groups, but admittedly fail to meet ideal ground truth standards. The clinically-annotated datasets that do exist are either proprietary or do not provide a clear mechanism for inquiring about availability. The dearth of large, shareable datasets based on actual clinical diagnoses and medical ground truth is problematic given recent research that calls into question the validity of proxy-based mental health annotations \cite{ernala2019methodological,harrigianmodels}. By leveraging privacy-preserving technology (e.g. blockchain, differential privacy) to share patient-generated data, researchers may ultimately be able to train more robust computational models \cite{elmisery2010privacy,zhu2016sharing,dwivedi2019decentralized}. In lieu of implementing complicated technical approaches to preserve the privacy of human subjects within mental health data, researchers may instead consider establishing secure computational environments that enable collaboration amongst authenticated users \cite{boebert1994data,rush2019icpsr}.

\textbf{Addressing Bias}. There remains more to be done to ensure models trained using these datasets perform consistently irrespective of population. Several studies in our review attempted to leverage demographically-matched or activity-based control groups as a comparison to individuals living with a mental health condition \cite{coppersmith-etal-2015-clpsych,Cohan2018SMHDAL}. A recent article found discrepancies between the prevalence of depression and PTSD as measured by the Centers for Disease Control and Prevention and as estimated using a model trained to detect the two conditions \cite{amir2019mental}. While the study posits reasons for the difference, it is unable to confirm any causal relationship. 

More recently, \citet{aguirre2021genderandracial} found evidence of demographic (gender and racial/ethnic) bias within datasets from \citet{coppersmith-etal-2014-quantifying, coppersmith2015quantifying} that can create fairness issues in downstream tasks. They found poor representation and strong group imbalance in these datasets; however, simple changes in dataset size and balance alone could not fully account for performance disparities between groups. Indeed, common signs of depression recognized in prior linguistic analyses (e.g. differences in distributions for some categories of LIWC) were found not to be equally informative for all demographics. Thus, while performance disparities between demographic groups may certainly arise due to poor representation at training time, disparities may also arise due to an ill-founded assumption that mental health outcomes for all groups can be treated equivalently \cite{kessler2003epidemiology,Choudhury2017GenderAC,shah2019predictive}. Either way, there exists a need to rethink dataset curation and model evaluation so traditionally underrepresented groups are not further hindered from receiving adequate mental health care.

This all said, the presence of downstream bias in mental health models is admittedly difficult to define and even more difficult to fully eliminate \cite{gonen2019lipstick,blodgett2020language}. Nonetheless, the lack of demographically-representative sampling described above would serve as a valuable starting point to address. Increasingly accurate demographic inference tools may aid in constructing datasets with demographically-representative cohorts \cite{huang2019hierarchical,wood2020using}. Researchers may also consider expanding the diversity of languages in their datasets to account for variation in mental health presentation that arises due to cultural differences \cite{Choudhury2017GenderAC,Loveys2018CrossculturalDI}. 

\bibliographystyle{acl_natbib}
\bibliography{clpsych}

\appendix




\section{Available Datasets}



Ultimately, we identified 35 unique mental health datasets that were available for distribution. A subset of annotations for these datasets, along with original reference information, can be found in Table \ref{tab:AvailableData} (see next page).  

We categorize dataset availability using four distinct distribution mechanisms.

\begin{itemize}
    \itemsep0em 
    \item \texttt{DUA}: The dataset requires researchers to sign a data usage agreement that outlines the terms and conditions by which the dataset may be analyzed; in some cases, this also requires institutional authorization and oversight (e.g. IRB approval)
    \item \texttt{API}: The dataset may be reproduced (with a reasonable degree of effort) using instructions provided in the dataset's primary article and access to a public-facing application programming interface (API)
    \item \texttt{AUTH}: The dataset may be accessed by directly contacting the original author(s)
    \item \texttt{FREE}: The dataset is hosted on a public-facing server, accessible by all without any additional restrictions
\end{itemize}

Of the datasets that were available for distribution via one of the above mechanisms, we noted the following 27 unique mental health conditions/predictive tasks: 

\begin{itemize}
    \itemsep0em 
    \item Attention Deficit Hyperactivity Disorder (ADHD)
    \item Alcoholism (ALC)
    \item Anxiety (ANX)
    \item Social Anxiety (ANXS)
    \item Asperger's (ASP)
    \item Autism (AUT)
    \item Bipolar Disorder (BI)
    \item Borderline Personality Disorder (BPD)
    \item Depression (DEP)
    \item Eating Disorder (EAT)
    \item Recovery from Eating Disorder (EATR)
    \item General Mental Health Disorder (MHGEN)
    \item Obsessive Compulsive Disorder (OCD)
    \item Opiate Addiction (OPAD)
    \item Opiate Usage (OPUS)
    \item Post Traumatic Stress Disorder (PTSD)
    \item Panic Disorder (PAN)
    \item Psychosis (PSY)
    \item Trauma from Rape (RS)
    \item Schizophrenia (SCHZ)
    \item Seasonal Affective Disorder (SAD)
    \item Self Harm (SH)
    \item Stress (STR)
    \item Stressor Subjects (STRS)
    \item Suicide Attempt (SA)
    \item Suicidal Ideation (SI)
    \item Trauma (TRA)
\end{itemize}

\begin{table*}[htb!]
\centering
\resizebox{\linewidth}{!}{%
\def\arraystretch{1.2}%
\begin{tabular}{lllllll}
\textbf{Reference} &    
    \textbf{Platform(s)} &    
    \textbf{Task(s)} &  
    \textbf{Level} & 
    \textbf{Individuals} & 
    \textbf{Documents} &          
    \textbf{Availability} \\ \hline \toprule
\citet{coppersmith-etal-2014-quantifying}  &
    Twitter &                  
    \begin{tabular}{@{}l@{}}BI, PTSD, SAD, \\ DEP\end{tabular}  & 
    Ind. &
    7k &
    16.7M &
    DUA \\ \hline
\citet{Coppersmith2014MeasuringPT}  &    
    Twitter &
    PTSD &
    Ind. &
    6.3k & 
    - & 
    DUA \\ \hline
\citet{Jashinsky2014TrackingSR}  & 
    Twitter & 
    SI &
    Doc. &
    594k &
    733k &
    API \\ \hline
\citet{lin2014user}  &    
    \begin{tabular}{@{}l@{}}Twitter,\\Sina Weibo,\\Tencent Weibo\end{tabular} &
    STR, STRS &
    Ind. &
    23.3k & 
    490k & 
    API \\ \hline
\citet{coppersmith-etal-2015-adhd}  & 
    Twitter & 
    \begin{tabular}{@{}l@{}}ANX, EAT, OCD,\\ SCHZ, SAD, BI, \\ PTSD, DEP, ADHD\end{tabular}   &     
    Ind. & 
    4k & 
    7M & 
    DUA \\ \hline
\citet{coppersmith-etal-2015-clpsych}  &
    Twitter &
    PTSD, DEP &
    Ind. &  
    1.7k &
    - &  
    DUA \\ \hline
\citet{de2015anorexia} &
    Tumblr &
    EAT, EATR &
    Ind. & 
    28k &
    87k &
    API \\ \hline
\citet{Kumar2015DetectingCI}  & 
    \begin{tabular}{@{}l@{}}Reddit,\\Wikipedia\end{tabular} &   
    SI &
    Ind. &
    66k &
    19.1k &
    API \\ \hline
\citet{mowery-etal-2015-towards}  &
    Twitter &
    DEP &
    Doc. & 
    - &
    129 &
    AUTH \\ \hline
\citet{chancellor2016recovery} &
    Tumblr &
    EATR &
    Ind. &  
    13.3k &
    67M &  
    API \\ \hline
\citet{coppersmith-etal-2016-exploratory}  &
    Twitter &
    SA &
    Ind. & 
    250 &
    - &  
    DUA \\ \hline
\citet{Choudhury2016DiscoveringST}  & 
    Reddit &
    \begin{tabular}{@{}l@{}}PSY, EAT, ANXS, \\ SH, BI, PTSD, \\ RS, DEP, PAN, \\ SI, TRA\end{tabular}   &  
    Ind. & 
    880 & 
    - &  
    API \\ \hline
\citet{gkotsis-etal-2016-language} &
    Reddit &
    \begin{tabular}{@{}l@{}}ANX, BPD, SCHZ, \\ SH, ALC, BI, \\ OPAD, ASP, SI, \\ AUT, OPUS\end{tabular}   &
    Ind. &
    - &  
    - &  
    API \\ \hline
\citet{lin2016does}  &
    Sina Weibo &
    STR &
    Doc. &  
    - &
    2.6k &  
    FREE \\ \hline
\citet{milne-etal-2016-clpsych} &
    Reach Out &
    SH &
    Doc. & 
    1.2k & 
    - & 
    DUA \\ \hline
\citet{Mowery2016TowardsAC}  &  
    Twitter &
    DEP & 
    Doc. & 
    - &   
    9.3k &
    AUTH \\ \hline
\citet{Bagroy2017ASM} &
    Reddit &
    MHGEN &
    Doc. &
    30k &
    43.5k &
    API \\ \hline
\citet{choudhury2017the} &
    Reddit &
    SI &
    Ind. &
    51k &
    103k &
    API \\ \hline
\citet{losada2017erisk} &
    Reddit &
    DEP &
    Ind. &
    887 &
    530k &
    DUA \\ \hline
\citet{saha2017modeling} &
    Reddit &
    STR &
    Doc. &
    - &
    2k &
    API \\ \hline
\citet{shen2017depression} &
    Twitter &
    DEP &
    Ind. &
    300M &
    10B &
    FREE \\ \hline
\citet{shen2017detecting} & 
    Reddit &
    ANX &
    Doc. &
    - &
    22.8k &
    API \\ \hline
\citet{Yates2017DepressionAS} &
    Reddit &
    DEP &
    Ind. & 
    116k &
    - &
    DUA \\ \hline
\citet{chancellor2018norms} &
    Reddit &
    EAT & 
    Doc. & 
    - &
    2.4M &
    API \\ \hline
\citet{Cohan2018SMHDAL} & 
    Reddit &
    \begin{tabular}{@{}l@{}}ANX, EAT, OCD, \\ SCHZ, BI, PTSD, \\ DEP, ADHD, AUT\end{tabular}   &
    Ind. & 
    350k & 
    - &
    DUA \\ \hline
\citet{dutta2018measuring} &
    Twitter &    
    ANX &    
    Ind. &
    200 & 
    209k &
    API \\ \hline
\citet{ireland2018within} &
    Reddit &    
    ANX &    
    Ind. &
    - & 
    - &
    API \\ \hline
\citet{Li2018TextBasedDA} &
    Reddit &
    MHGEN &
    Ind. &
    1.8k &
    - & 
    API \\ \hline
\citet{losada2018overview} &
    Reddit &    
    EAT, DEP &    
    Ind. &
    1.5k & 
    1.2M &
    DUA \\ \hline
\citet{pirina2018identifying} &
    Reddit &    
    DEP &    
    Doc. &
    - & 
    1.2k &
    API \\ \hline
\citet{Shing2018ExpertCA} &
    Reddit &    
    SI &    
    Ind. &
    1.9k & 
    - &
    DUA \\ \hline
\citet{sekulic2018not} &
    Reddit &
    BI &
    Ind. &
    7.4k &
    - &
    API \\ \hline  
\citet{wolohan-etal-2018-detecting} &
    Reddit &
    DEP & 
    Ind. & 
    12.1k &
    - &
    API \\ \hline
\citet{turcan-mckeown-2019-dreaddit} & 
    Reddit &
    STR &
    Doc. &
    - &
    2.9k &
    FREE \\ \hline
\citet{zirikly-etal-2019-clpsych} &
    Reddit &
    SI &
    Ind. &
    496 & 
    32k &
    DUA \\
\end{tabular}
}
\caption{Characteristics of datasets that meet our inclusion criteria and are known to be accessible. The full set of annotations may be found in our digital directory (\url{https://github.com/kharrigian/mental-health-datasets}).}  \label{tab:AvailableData}
\end{table*}

\end{document}